\documentclass[10pt,twocolumn,letterpaper]{article}

\usepackage{cvpr}
\usepackage{times}
\usepackage{epsfig}
\usepackage{graphicx}
\usepackage{amsmath}
\usepackage{amssymb}
\usepackage{amsthm}
\usepackage{algpseudocode,algorithm,algorithmicx}

\usepackage{stmaryrd}
\usepackage{wrapfig}
\usepackage{multirow}

\usepackage{array}
\makeatletter
\newcommand{\thickhline}{%
	\noalign {\ifnum 0=`}\fi \hrule height 1pt
	\futurelet \reserved@a \@xhline
}
\newcolumntype{"}{@{\hskip\tabcolsep\vrule width 1pt\hskip\tabcolsep}}
\makeatother

\usepackage[pagebackref=true,breaklinks=true,letterpaper=true,colorlinks,bookmarks=false]{hyperref}

 \cvprfinalcopy 

\def\cvprPaperID{1880} 

\ifcvprfinal\fi
\begin{document}
	
	\title{Deep Cocktail Network: \\ Multi-source Unsupervised Domain Adaptation with Category Shift}
	
	\author{Ruijia Xu$^{1,\dagger}$, \ Ziliang Chen$^{1,\dagger}$, \ Wangmeng Zuo$^{2}$, \ Junjie Yan$^{3}$, \ Liang Lin$^{1,3}$\\ $^1$Sun Yat-sen University  \ \ $^2$Harbin Institute of Technology \ \  $^3$SenseTime Group Limited \\
		\tt\small xurj3@mail2.sysu.edu.cn, c.ziliang@yahoo.com, \tt\small wmzuo@hit.edu.cn, \\ \tt\small yanjunjie@sensetime.com, \tt\small linliang@ieee.org \thanks{$\dagger$ indicates equal contribution. ∗Corresponding author: Liang Lin}
	}
	
	\maketitle
	
	\begin{abstract}
		
	 Unsupervised domain adaptation (UDA) conventionally assumes labeled source samples coming from a single underlying source distribution. Whereas in practical scenario, labeled data are typically collected from diverse sources. The multiple sources are different not only from the target but also from each other, thus, domain adaptater should not be modeled in the same way. Moreover, those sources may not completely share their categories, which further brings a new transfer challenge called category shift. In this paper, we propose a deep cocktail network (DCTN) to battle the domain and category shifts among multiple sources. Motivated by the theoretical results in \cite{mansour2009domain}, the target distribution can be represented as the weighted combination of source distributions, and, the multi-source unsupervised domain adaptation via DCTN is then performed as two alternating steps: i) It deploys multi-way adversarial learning to minimize the discrepancy between the target and each of the multiple source domains, which also obtains the source-specific perplexity scores to denote the possibilities that a target sample belongs to different source domains. ii) The multi-source category classifiers are integrated with the perplexity scores to classify target sample, and the pseudo-labeled target samples together with source samples are utilized to update the multi-source category classifier and the feature extractor. We evaluate DCTN in three domain adaptation benchmarks, which clearly demonstrate the superiority of our framework.

	\end{abstract}
	
	\begin{center}
		\begin{figure}[t] \centering
			\includegraphics[width=3.2in]{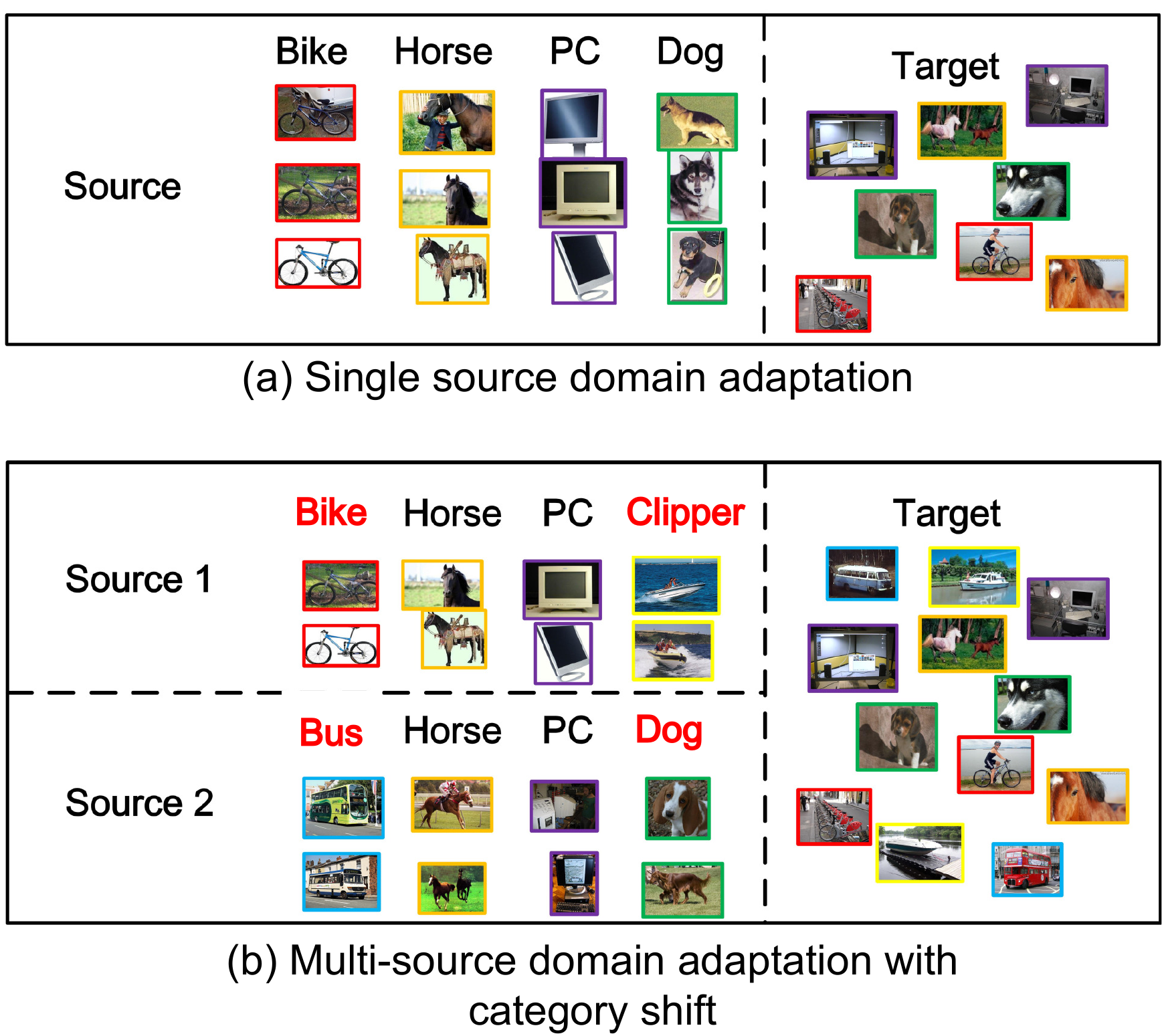}
			\caption{(a). Single source domain adaptation (UDA) assumes that source samples are drawn from some underlying distribution under the i.i.d. condition. (b). Multiple source domain adaptation (MDA) assume source data are collected from different source distributions. Category shift is a new protocol in MDA, where domain shift and categorical disalignment co-exist among the sources.}\label{p}
		\end{figure}
	\end{center}
	
	\section{Introduction}
	Recent advances in deep learning have significantly improved the state-of-the-arts across a variety of visual learning tasks \cite{krizhevsky2012imagenet} \cite{ren2015faster} \cite{long2015fully} \cite{cao2016realtime} \cite{xu2015show}.
	These achievements, to a great extent, should be attributed to the availability of large scale labeled data for supervised learning.
	When turning to (Unsupervised) domain adaptation (UDA, see Fig. \ref{p}(a)) \cite{pan2010survey} \cite{pan2011domain} \cite{gong2012geodesic}, we do not have the labels of the data in target domain, but have massive labeled data in source domain.
	One natural solution is to learn a deep model on the labeled source data and deploy it to target domain.
	However, due to the presence of \emph{domain shift} \cite{gretton2009covariate}, the performance of the learned model tends to degrade heavily in the target domain.
	To mitigate the model damage caused by the domain shift, UDA learns to map the data from  both domains into a common feature space by minimizing domain distribution discrepancy, the source classifier can then be directly applied to target instances.
	While early UDA studies mainly focus on shallow models \cite{pan2011domain} \cite{gong2012geodesic}, with the dramatic upsurge of convolutional neural networks (CNNs), deep UDA has emerged as a thriving solution and has achieved many stunning results \cite{hoffman2017simultaneous} \cite{bousmalis2016unsupervised} \cite{gebru2017fine}.

	
	However, most existing deep UDA methods assume that there is only a single source domain and the labeled source data are implicitly sampled from a same underlying distribution.
	In practice, it is very likely that we have multiple source domains.
	For example, when training object recognition models for household robots, one can exploit the labeled images either from Amazon.com (Source 1) or Flickr (Source 2).
	Moreover, the large scale dataset, e.g., \emph{ImageNet} \cite{deng2009imagenet} may be built upon diverse sources from the Internet, and is inappropriate to be treated as a single domain in UDA.
	Consequently, multiple source unsupervised domain adaptation (MDA) is both feasible in practice and more valuable in performance, and has received considerable attention in application fields \cite{yang2007cross}\cite{duan2012domain}\cite{jhuo2012robust} \cite{liu2016structure}.

	Despite the rapid progress in deep UDA, seldom studies have been given to deep MDA which is much more challenging due to the following reasons.
	Firstly, with possible domain shifts among sources, it's improper to apply single source UDA via combining all source domains.
	Secondly, different source domains convey complimentary information to target domain.
	Based on Liebig's law of the minimum, it is too strict to eliminate the distribution discrepancy between target domain and each source domain, and may be harmful to the model performance.
	Finally, as illustrated in Fig.\ref{p}(b), different source domains may not completely share their categories (i.e., \emph{category shift}), some category of samples may appear in one source domain but not in another.
	MDA should take both category shift and domain shift into account, and is thus more challenging to handle.
	%


	In this paper, we propose the \emph{deep cocktail network} (DCTN) for MDA.
	Inspired by the \emph{distribution weighted combining rule} in \cite{mansour2009domain}, the target distribution can be represented as the weighted combination of the multi-source distributions.
	Suppose the classifier for each source domain is known.
	An ideal target predictor can be obtained by integrating all source predictions based on the corresponding source distribution weights.
	Therefore, besides of the feature extractor, DCTN also includes a (multi-source) category classifier to predict the class from different sources, and a (multi-source) domain discriminator to produce multiple source-target-specific perplexity scores as the approximation of source distribution weights.
	Analogous to make cocktails, the multi-source class predictions are integrated with the perplexity scores to classify the target sample, and thus the proposed method is dubbed by deep cocktail network (DCTN).

	During training, the learning algorithm for DCTN performs the following two alternating adaptation steps:
	(i) the domain discriminator is updated by using multi-way adversarial learning to minimize the domain discrepancies between target and each source, then to predict multi-source perplexity scores;
	(ii) the feature extractor and the category classifier are discriminatively fine-tuned with multi-source labeled and target pseudo-labeled data.
	The multi-way adversarial adaptation implicitly reduces domain shifts among those sources. 
	The discriminative adaptation helps to learn more classifiable features \cite{saito2017asymmetric}, and partially prevents the \emph{negative transfer} \cite{pan2010survey} from the mis-matching categories.
	Empirical studies on three domain adaptation benchmarks also demonstrate the effectiveness of our DCTN framework.

	Our work contributes in the three aspects: \textbf{1)} We present a novel and realistic MDA protocol termed \emph{category shift} that relaxes the requirement on the shared category set among any source domains.
	\textbf{2)} Inspired from the distribution weighted combining rule, we proposed the \emph{deep cocktail network} (DCTN) together with the alternating adaptation algorithm to learn transferable and discriminative representation.
	\textbf{3)} We conduct comprehensive experiments on three well-known benchmarks, and testify our model in both the vanilla and the \emph{category shift} settings. Our method has achieved the state of the art across most transfer tasks. 
	\section{Related Work}
	
	\textbf{Unsupervised domain adaptation with single source.} Provided a source domain with ground truth and target domain without labels, unsupervised domain adaptation (UDA) aims at learning a model well-performing on target distribution. Since the source and the target belong to different distributions, the technical problem in UDA is how to reduce the domain shift across the source and the target. Inspired by the two-sample test \cite{gretton2007kernel}, domain discrepancy based methods, e.g., shallow-model-based TCA \cite{pan2011domain}, JDA \cite{baktashmotlagh2016distribution}; deep-model-based DAN \cite{long2015learning}, WMMD \cite{yan2017mind}, RTN \cite{long2016unsupervised}, leverage different distribution measures as domain regularizer to attain domain-invariant feature. Adversarial learning behaves effective to learn more transferable representations. It defines a couple of networks and trains them in the opposite direction: a domain discriminator minimizes the classification error to distinguish samples from source and target, while domain mapping learns transferable representations indistinguishable by the domain discriminator. Recent relevant researches perform superior in visual recognition cross domain \cite{long2016unsupervised} \cite{gebru2017fine} and task \cite{motiian2017few} and transfer structure learning \cite{bousmalis2016unsupervised} \cite{hoffman2016fcns}. Besides of these two mainstreams, there are diverse methods to learn domain-invariant features: semi-supervised method \cite{saito2017asymmetric}, domain reconstruction \cite{ghifary2016deep}, duality \cite{haeusser2017associative}, alignments \cite{fernando2013unsupervised} \cite{zhang2017joint} \cite{sun2016return}, manifold learning \cite{gong2012geodesic}, tensor methods \cite{koniusz2016domain}\cite{lu2017unsupervised}, etc.
	
	\textbf{Domain adaptation with multiple sources.} The UDA methods mentioned above mainly consider target \emph{vs.} single source. If multiple sources are available, the domain shift among sources should also be account for. The research originates from A-SVM \cite{yang2007cross} that leverages the ensemble of source-specific classifiers to tune the target categorization model, and there have been a variety of shallow models invented to tackle the MDA problem \cite{duan2012domain} \cite{jhuo2012robust} \cite{liu2016structure}. MDA also develops with theoretical supports \cite{blitzer2008learning} \cite{ben2010theory} \cite{mansour2009domain}. Blitzer et al \cite{blitzer2008learning} provides the first learning bound for MDA. Mansour et al \cite{mansour2009domain} claims that an ideal target hypothesis can be represented by a distribution weighted combination of source hypotheses. This methodology termed \emph{ distribution weighted combining rule}, closely means that, if the relations between target and each source can be discovered, we are able to use multiple source-specific classifiers to obtain an ideal target class prediction.
	
	\textbf{Continual transfer learning, domain generalization.} There are two branches of transfer learning closely relate to MDA. The first is continual transfer learning (CTL) \cite{shin2017continual} \cite{rebuffi2017learning}. Similar to continual learning \cite{kirkpatrick2016overcoming}, CTLs train the learner to sequentially master multiple tasks across multiple domains. The second is domain generalization (DG) \cite{ghifary2015domain} \cite{motiian2017unified}, which solely uses the existing multiple labeled domains for training regardless of the unlabeled target samples. Both of the problems are solved by supervised learning approaches, and distinguished from MDA with unlabeled training samples.  
	
	
	


	
	\section{Problem Setup}
	
	\textbf{Vanilla MDA.} In the context of multi-source transfer, there are $N$ different underlying source distributions denoted as $\{p_{s_j}(x,y)\}^N_{j=1}$. The labeled source domain images \begin{small}$\{(X_{s_j},Y_{s_j})\}^N_{j=1}$\end{small} are drawn from those distributions respectively, where \begin{small}
		$X_{s_j}= \{x^{s_j}_i\}^{|X_{s_j}|}_{i=1}$\end{small} represents images from source 
		$j$ and \begin{small}
	$Y_{s_j}=\{y^{s_j}_i\}^{|Y_{s_j}|}_{i=1}$
\end{small} is the corresponding ground-truth set. Besides, we have target distribution \begin{small}
$p_t(x,y)$
\end{small}, from which target image set \begin{small}
$X_t = \{x^t_i\}^{|X_t|}_{i=1}$
\end{small} are sampled yet without label observation \begin{small}
$Y_t$
\end{small}. Those $N$+1 datasets have been treated as an training set ensemble, and the test set \begin{small}
$(X_{test}, Y_{test}) = \{x^{test}_i, y^{text}_i\}^{|X_{test}|}_{i=1}$
\end{small} are drawn in target distribution to evaluate the model adaptation performance.


\begin{figure*}[t]
	\centering
	\includegraphics[width=6in,height=2.4in]{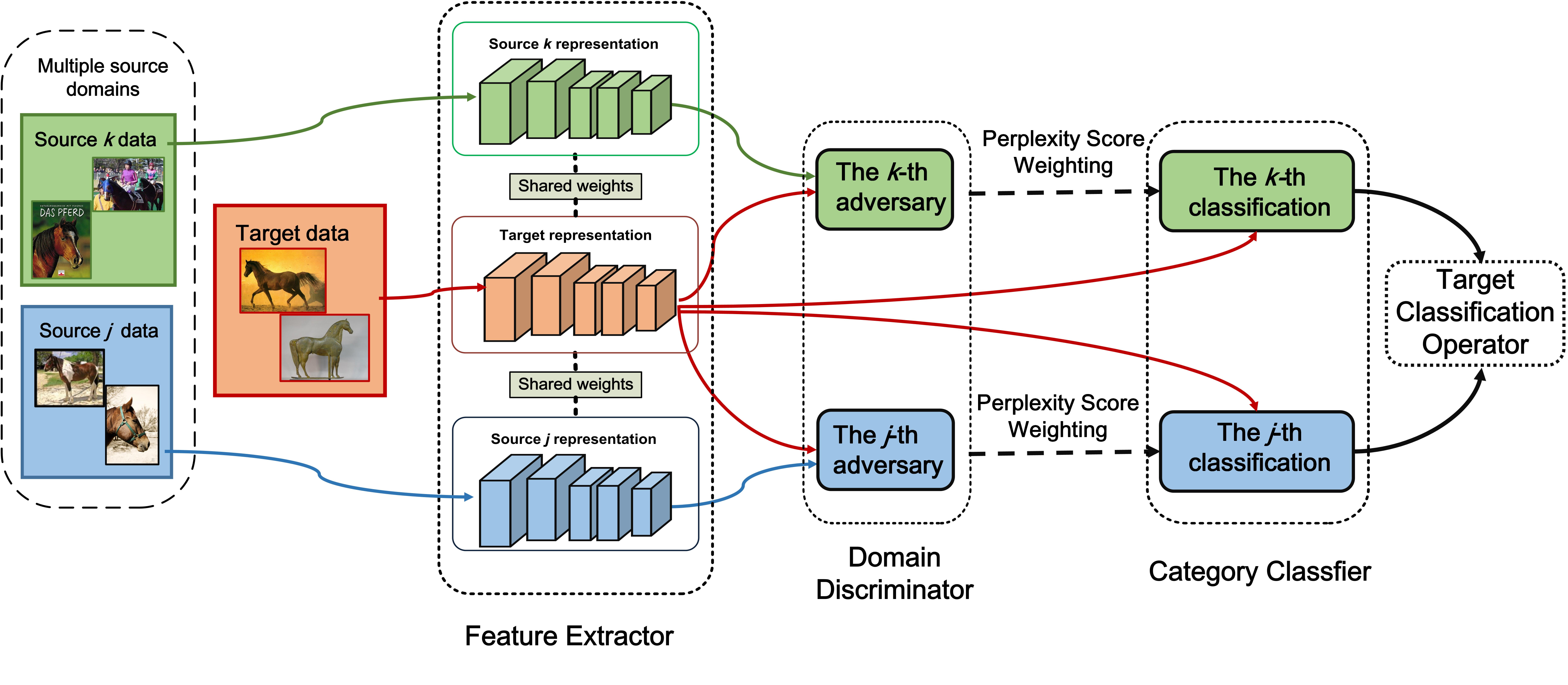}
	\caption{An overview of the proposed Deep Cocktail Network (DCTN). Our framework receives multi-source instances with annotated ground truth and adapts to classify the target samples. Let's consider the source $j$ and $k$ for simplicity. \textbf{i)} The feature extractor maps target, source $j$ and $k$ into a common feature space. \textbf{ii)} The category classifier receives target feature and produces the $j$-th and $k$-th classifications based upon the categories in source $j$ and $k$ respectively. \textbf{iii)} The domain discriminator receives features from source $j$, $k$ and target, then offers the $k$-th advesary between target and source $k$, as well as the $j$-th advesary between target and source $j$. The $j$-th and $k$-th advesary provide source $j$ and $k$ perplexity scores to weight the $j$-th and $k$-th classifications correspondingly. \textbf{iv)} The target classification operator integrates all weighted classification results then predicts the target class across category shifts. Best viewed in color. }\label{pic}
\end{figure*}

\textbf{Category Shift.}
Under the vanilla MDA setting, samples from diverse sources share a same category set. In contrast to this old fashion, we introduce a new MDA protocol where the categories from different sources might be also different. Formally speaking, given a category set $\mathcal{C}_{s}=\overset{|Y_s|}{\underset{i=1}{\bigcup}}\{y^{s}_{i}\}$ as a class set of $Y_{s}$ for domain $s$, the relation between $\mathcal{C}_{s_{j_1}}$ and $\mathcal{C}_{s_{j_2}}$ has been generalized from \begin{small}
	$\mathcal{C}_{s_{j_1}}\cup\mathcal{C}_{s_{j_2}} = \mathcal{C}_{s_{j_1}}\cap\mathcal{C}_{s_{j_2}}$
\end{small} to \begin{small}
$\mathcal{C}_{s_{j_1}}\cap\mathcal{C}_{s_{j_2}} \subseteq \mathcal{C}_{s_{j_1}}\cup\mathcal{C}_{s_{j_2}}$
\end{small}, where \begin{small}
$\mathcal{C}_{s_{j_1}}\cap\mathcal{C}_{s_{j_2}}$
\end{small} denotes public classes between sources $j_1$ and $j_2$. Let target domain get labeled by the union of all categories in those sources (\begin{small}
$\mathcal{C}_{t}=\overset{M}{\underset{j=1}{\bigcup}}\mathcal{C}_{s_{j}}$
\end{small}), then we term $\mathcal{C}_{s_{j_1}}\cap\mathcal{C}_{s_{j_2}} \neq \mathcal{C}_{s_{j_1}}\cup\mathcal{C}_{s_{j_2}}$ as \emph{category shift} in multiple source domains \begin{small}$\{(X_{s_j},Y_{s_j})\}^N_{j=1}$\end{small}. 

\textbf{Compared with Open Set DA.} Open set domain adaptation \cite{busto2017open} is a new single-source transfer protocol, where the classes between the source and the target domains are allowed to be different. The uncommon classes are unified as a negative category called ``unknown''. In contrast, category shift consider the specific disaligned categories among multiple sources to enrich the classification in transfer. In fact, the open set DA can also be developed to our category shift setting, where the unshared classes are viewed unobservable. Such study will be investigated in our future work. 


\section{Deep Cocktail Network}
Irrespective of either vanilla or category shift scenarios, MDAs are challenging to tackle. In this section, we introduce \emph{deep cocktail network} (DCTN), an adversarial domain adaptation framework for both MDA protocols. It connects to the \emph{distribution weighted combining rule} \cite{mansour2009domain}, and what's more, can be easily transplanted to suit the shifted categories without model reconfiguration.

\subsection{Architecture}
Our framework consists of four components: three subnets, i.e., \emph{feature extractor}, \emph{(multi-source) domain discriminator}, \emph{(multi-source) category classifier}, and a non-learnable \emph{target classification operator}, as shown in Fig.\ref{pic}.

\textbf{Feature extractor} $F$ incorporates deep convolution nets as the backbone, and is supposed to map all images from $N$ sources and target into a common feature space. We employ adversarial learning to obtain the optimal mapping, because it can successfully learn both domain-invariant features and each target-source-specific relations.

\textbf{(Multi-source) domain discriminator} $D$ is built upon $N$ source-specific discriminators $\{D_{s_j}\}^N_{j=1}$ for adversary. Given image $x$ from the source $j$ or the target domain, the domain discriminator $D$ receives the features $F(x)$, then the source-specific discriminator $D_{s_j}$ classifies whether $F(x)$ originates from the source $j$ or the target. The data flow from source $j$ doesn't trigger other source discriminators, yet for the data flow from each target instance $x^t$, the domain discriminator $D$ yields the $N$ source-specific discriminative results $\{D_{s_j}(F(x^t))\}^N_{j=1}$. They are used to update the domain discriminator $D$, also to supply the target-source perplexity scores $\{\mathcal{S}_{cf}(x^t;F,D_{s_j})\}^N_{j=1}$ to the target classification operator
\begin{small}
	\begin{equation}
	\begin{aligned}
	\mathcal{S}_{cf}(x^t;F,D_{s_j}) = -\log(1- D_{s_j}(F(x^t)))+ \alpha_{s_j}\label{sc}
	\end{aligned}
	\end{equation}
\end{small}where $\alpha_{s_j}$ is the \emph{source-specific concentration constant}. It is obtained by averaging the source $j$ discriminator losses over $X_{s_j}$.

\textbf{(Multi-source) category classifier} $C$ is a multi-output net composed by $N$ source-specific predictors $\{C_{s_j}\}^N_{j=1}$. Each predictor $C_{s_j}$ is a softmax classifier configured by the category set in the corresponding source $j$. The category classifier takes an image mapping as input, then for the image from source $j$, only the value from $C_{s_j}$ get activated and provides the gradient for training. For a target image $x^t$ instead, all source-specific predictors provide $N$ categorization results $\{C_{s_j}(F(x^t))\}^N_{j=1}$ to the target classification operator.

\textbf{Target classification operator} is the key to classify target samples. In specific, for each target feature $F(x^t)$, the target classification operator takes each source perplexity score $\mathcal{S}_{cf}(x^t;F,D_{s_j})$ to re-weight the corresponding source-specific prediction $C_{s_j}(F(x^t))$, then accumulates the results to classify target $x^t$. If the class $c\in \underset{j=1}{\overset{N}{\bigcup}} \{\mathcal{C}_{s_j}\}$ is considered, the confidence $x^t$ belongs to $c$ presents as

\begin{small}
	\begin{equation}
		\begin{aligned}
			Confidence(c|x^t) &:= \\ \sum_{c\in\mathcal{C}_{s_j}}\frac{\mathcal{S}_{cf}(x^t;F,D_{s_j})}{{\underset{c\in\mathcal{C}_{s_k}}{\sum}}\mathcal{S}_{cf}(x^t;F,D_{s_k})} \ &C_{s_j}(c|F(x^t))\label{k1}
		\end{aligned}
	\end{equation}
\end{small}where $C_{s_j}(c|F(x^t))$ denotes the softmax value of source $j$ corresponding to class $c$. $x^t$ is categorized into the class with the highest confidence. The sum $\underset{c\in\mathcal{C}_{s_j}}{\sum}$ means only those sources with class $c$ can join the perplexity score weighting. It's invented to incorporate both the vanilla and the \emph{category
	shift} settings. Since the module independently estimates each class confidence, the variation in shifting categories merely modifies the class combination in the target classification operator, but not the structures or the parameters in the three subnets.

\textbf{Connection to distribution weighted combining rule.} Let $\{\mathcal{D}_{s_j}\}^N_{j=1}$ and $\mathcal{D}_t$ denote sources and target distributions\footnote{Since each sample $x$ corresponds to an unique class $y$, $\{\mathcal{D}_{s_j}\}^N_{j=1}$ and $\mathcal{D}_t$ can be viewed as an equivalent embedding from $\{p_{s_j}(x,y)\}^N_{j=1}$ and $p_{t}(x,y)$ that we have discussed. }, and given an instance $x$, $\{\mathcal{D}_{s_j}(x)\}^N_{j=1}$ and $\mathcal{D}_t(x)$ denote the probabilities that $x$ is generated from $\{\mathcal{D}_{s_j}\}^N_{j=1}$ and $\mathcal{D}_t$, respectively. In the \emph{distribution weighted combining rule} \cite{mansour2009domain}, the target distribution is treated as a mixture of the multi-source distributions with the coeffients by normalized source distributions weighted by unknown positive $\{\lambda_j\}^N_{j=1}$, namely $\mathcal{D}_t(x) = \overset{N}{\underset{c\in\mathcal{C}_{s_k}}{\sum}}\lambda_k\mathcal{D}_{s_k}(x)$. The ideal target classifier $C_t(c|x^t)$ presents as the weighted combination of source classifiers $\{C_{s_j}(c|F(x^t))\}^M_{j=1}$:\begin{small}
	\begin{equation}
	\begin{aligned}
	C_t(c|x^t) =  \sum_{c\in\mathcal{C}_{s_j}}\frac{\lambda_j\mathcal{D}_{s_j}(x^t)}{\underset{c\in\mathcal{C}_{s_k}}{\sum}\lambda_k\mathcal{D}_{s_k}(x^t)} \ C_{s_j}(c|F(x^t))
	\end{aligned}\label{k2}
	\end{equation}
\end{small}Note a fact that, with the increase of the probability that $x^t$ from source $j$, $x^t$ becomes similar to the sample from source $j$. It holds \begin{small}
$D_{s_j}(F(x^t))\rightarrow 1$
\end{small} and results in \begin{small}$-\log(1- D_{s_j}(F(x^t)))$\end{small} increasing. Hence it maintains $\lambda_j\mathcal{D}_{s_j}(x^t)\propto \mathcal{S}_{cf}(x^t;F,D_{s_j})$ in the multiple source domains. Replace the source distributions with the normalized source perplexity scores, then $C_t(c|x^t)$ corresponds to the target classification operator in Eq.\ref{k1}. The formula physically implies that target images should be categorized by the classifiers from multiple sources, with whose features more similar to target, the source classifiers' prediction are more trustful.

\subsection{Learning}
Our framework admits an alternative adaptation pipeline. Briefly, after a proper pre-training, DCTN employs a multi-way adversary to acquire a mutual mapping from all domains, then further, the feature extractor and the category classifier are trained with multiple sources labeled and target pseudo-labeled images. The two stages repeat until the maximal epoch is reached.

\textbf{Pre-training} Pre-trained feature extractor and category classifier are the prerequisites for the alternative process. At the very start, we take all source images to jointly train the feature extractor $F$ and the category classifier $C$. Those networks and the target classification operator then predict categories for all target images\footnote{Since the domain discriminator hasn't been trained, we take the uniform distribution simplex weight as the perplexity scores to the target classification operator.} and annotate those with high confidences. Finally, we obtain the pre-trained feature extractor and category classifier via further fine-tuning them with sources and the pseudo-labeled target images. The alternative paradigm begins after this pretraining.

\subsubsection{Multi-way Adversarial Adaptation}
\ \ \ \ Our first stage multi-source domain adaptation are now described as follow:

\begin{small}
	\begin{equation}
	\underset{F}{\min} \ \underset{D}{\max} \ V(F,D;\overline{C}) = \mathcal{L}_{adv}(F,D) + \mathcal{L}_{cls}(F,\overline{C})\label{adv}
	\end{equation}\end{small}where
\begin{small}
	\begin{equation}
	\begin{aligned}
	\mathcal{L}_{adv}(F, \ &D)  = \frac{1}{N}\sum_{j}^{N}\mathbb{E}_{x\sim X_{s_j}}[\log D_{s_j}(F(x))] \\ &+ \mathbb{E}_{x^t\sim X_{t}}[\log(1- D_{s_j}(F(x^t)))]
	\end{aligned}
	\end{equation}\end{small}where the first term denotes our adversarial mechanism, and the second term is a multi-source classification losses. The classifier $C$ is fixed as $\overline{C}$ to provide stable gradient values.

The optimization based on Eq.\ref{adv} works well for $D$ but not $F$. Since the feature extractor learns the mapping from the multiple sources and the target, the domain distributions become simultaneously changing in adversary, which results in an oscillation then spoils our feature extractor. Towards such concern, Tzeng et al.\cite{tzeng2017adversarial} mentioned when source and target feature mappings share their architectures, the domain confusion can be introduced to replace the adversarial objective, which performs stable to learn the mapping $F$. Extend it to our scenario, we have the following multi-domain confusion loss:

\begin{small}
	\begin{equation}
	\begin{aligned}
	\mathcal{L}_{adv}(F,\ &D)  =  \frac{1}{N}\sum_{j}^{N} \mathbb{E}_{x\sim X_{s_j}} \mathcal{L}_{cf}(x;F,D_{s_j})  \\ + \   &\mathbb{E}_{x\sim X_{t}} \mathcal{L}_{cf}(x^t;F,D_{s_j})\label{confusion}
	\end{aligned}
	\end{equation}
\end{small}where \begin{small}\begin{equation}\begin{aligned}
\mathcal{L}_{cf}(x;F,&D_{s_j}) = \\\frac{1}{2}\log D_{s_j}(F(x)) + \frac{1}{2}&\log(1- D_{s_j}(F(x)))
\end{aligned}
\end{equation}
\end{small}

\begin{algorithm}[t]			
	\caption{Mini-batch Learning via online hard domain batch mining} \label{A2}
	\begin{small}
		\hspace*{0.02in}{\bf Input:} Mini-batch $\{x^t_i,\{x^{s_j}_i,y^{s_j}_i\}^N_{j=1}\}^M_{i=1}$ sampled from $X_t$ and $\{(X_{s_j},Y_{s_j})\}^N_{j=1}$ respectively; feature extractor $F$; domain discriminator $D$; category classifier $\overline{C}$.\\
		\hspace*{0.02in}{\bf Output:} Updated $F'$.
		\begin{algorithmic}[1]	
			\State Select the source domain $j^\ast\in[N]$, where \begin{small}
				$j^\ast = \overset{N}{\underset{j}{\arg\max}}\{\sum_{i}^{M}-\log D_{s_j}(F(x_i^{s_j}))-\log (1-D_{s_j}(F(x_i^{t})))\}^N_{j=1}$;
			\end{small}
			\State $\mathcal{L}_{adv}^{s_{j^\ast}}  =  \sum_{i}^{M}\mathcal{L}_{cf}(x^{s_{j^\ast}}_i;F,D_{s_{j^\ast}}) + \mathcal{L}_{cf}(x_i^t;F,D_{s_{j^\ast}})$
			\State Replace $\mathcal{L}_{adv}$ in Eq.\ref{adv} with $\mathcal{L}_{adv}^{s_{j^\ast}}$, update $F$ by Eq.\ref{adv}.\\
			\Return $F' = F$.
		\end{algorithmic}  \end{small}
	\end{algorithm}

	\textbf{Online hard domain batch mining} In the stochastic gradient manner, the multi-way adversarial learning receive $N$ samples from $N$ sources respectively to update $F$ in each iteration. However, the samples from different sources are sometimes useless to improve the adaptation to the target, and as the training proceeds, more redundant source samples turn to draw back the whole model performance. To mitigate this negative effect, we proposed a simple yet effective multi-source batch mining technique to improve the training. For a specific target batch $\{x^t_i\}^M_{i=1}$, we consider $N$ sources batches $\{\{x_i^{s_1}\}^M_{i=1},\cdots,\{x_i^{s_N}\}^M_{i=1}\}$. Each source-target discriminator loss \begin{small}
		$\{\sum_{i}^{M}-\log D_{s_j}(F(x_i^{s_j}))-\log (1-D_{s_j}(F(x_i^{t})))\}^N_{j=1}$
	\end{small}, is viewed as the degrees to distinguish $x^t_i$ from $N$ source samples. Hence $F$ performs worst to transform the target samples to confuse source $j^\ast$, which results in \begin{small}
	$j^\ast = \overset{N}{\underset{j}{\arg\max}}\{\sum_{i}^{M}-\log D_{s_j}(F(x_i^{s_j}))-\log (1-D_{s_j}(F(x_i^{t})))\}^N_{j=1}$
\end{small}. Based upon the domain confusion loss, we use the source $j^\ast$ and the target samples in the mini-batch to train the feature extractor. This stochastic learning method is represented by the Algorithm.\ref{A2}.

\subsubsection{Target Discriminative Adaptation}
\ \ \ \ Aided by the multi-way adversary, DCTN has been able to obtain good domain-invariant features, yet not surely classifiable in the target domain. David et al \cite{ben2010theory} demonstrates that, to apply source classifier in the target domain, it must acquiesces in a classifier that works well on both the domains. However, in the MDA setting, such ideal across-domain classifier must account for the non-consistency among different sources, even with their shifting categories. It's obvious that such MDA-based classifier is too difficult to access.

To further approach an ideal target classifier, we directly incorporate target samples to learn discriminative features with multiple sources. We propose an auto-labeling strategy to annotate target samples, then jointly train our feature extractor and multi-source category classifier with source and target images by their (pseudo-) labels. Hence, the discriminative adaptation of DCTN presents as

\begin{small}
	\begin{equation}
	\begin{aligned}
	\underset{F ,\ C}{\min} \ \ \mathcal{L}_{cls}(F, C) = \sum_{j}^{N} \ \mathbb{E}_{(x,y)\sim (X_{s_j},Y_{s_j})}[\mathcal{L}(&C_{s_j}(F(x)),y)]  \\
	+ \ \ \mathbb{E}_{(x^t,\hat{y})\sim (X^p_{t},Y^p_{t})}[\sum_{\hat{y}\in \mathcal{C}_{\hat{s}}}&\mathcal{L}(C_{\hat{s}}(F(x^t)),\hat{y})] \label{cls}
	\end{aligned}\end{equation}\end{small}where the first and second terms denote the classification losses from multiple source images $\{X_{s_j},Y_{s_j}\}^N_{j=1}$, and target images with pseudo labels $\{X^P_{t},Y^P_{t}\}$ respectively. We apply the target classification operator to assign pseudo labels, and the samples with the confidence higher than a preseted threshold $\gamma$ will be selected into $X^P_{t}$.

Since the target predictions come from the integration of multi-source predictions, there is no explicit learnable target classifier. As illustrated in the second term of Eq.\ref{cls}, we apply the multi-source category classifier to back-propagate pseudo target classification errors. Concretely, given a target instance $x^t$ with pseudo-labeled class $\hat{y}$, we find those sources $\hat{s}$ include this class ($\hat{y}\in \mathcal{C}_{\hat{s}}$), then update our network via the sum of the multi-source classification losses, namely, $\sum_{\hat{y}\in \mathcal{C}_{\hat{s}}}\mathcal{L}(C_{\hat{s}}(F(x^t)),\hat{y})$ in the second term.

The alternative adaptation pipline of DCTN has been summarized in Algorithm.\ref{A1}.

\begin{algorithm}[t]			
	\caption{Learning algorithm for DCTN} \label{A1}
	\begin{small}
		\hspace*{0.02in}{\bf Input:} $N$ source labeled datasets $\{X_{s_j},Y_{s_j}\}^N_{j=1}$; target unlabeled dataset $X_t$; initiated feature extractor $F$; category classifier $C$ and domain discriminator $D$; confidence threshold $\gamma$; adversarial iteration threshold $\beta$.\\
		\hspace*{0.02in}{\bf Output:} well-trained feature extractor $F^\ast$, domain discriminator $D^\ast$ and category classifier $C^\ast$.
		\begin{algorithmic}[1]
			\State \textbf{Pre-train $C$ and $F$}
			\While{not converged}
			\State \textbf{Multi-way Adversarial Adaptation:}
			\For{1:$\beta$}
			\State Sample mini-batch from $\{X_{s_j}\}^N_{j=1}$ and $X_t$;
			\State Update $D$ by Eq.\ref{adv};
			\State Update $F$ by Algorithm.\ref{A2};sequentially
			\EndFor
			\State \textbf{Target Discriminative Adaptation:}
			\State Estimate confidence for $X_t$ by Eq.\ref{k1} with similarities offered by Eq.\ref{sc}. Samples $X^P_t\subset X_t$ with confidence larger than $\gamma$ get annotations $Y^P_T$;
			\State Update $F$ and $C$ by Eq.\ref{cls}.
			\EndWhile \\
			\Return $F^\ast = F; C^\ast = C; D^\ast = D$.
		\end{algorithmic}
	\end{small}
\end{algorithm}

\section{Experiments}
In the context of MDA for visual classification, we evaluate the accuracy of the predictions from the target classification operator in all experiments, and both of the vanilla setting and the category shift have been validated. Our DCTN are all implemented in the PyTorch\footnote{http://pytorch.org/} platform. We report the major results in the paper, and more implementation information and results have been detailed in the Appendix.

\subsection{Benchmarks} Three widely used UDA benchmarks \emph{Office-31} \cite{saenko2010adapting}, \emph{ImageCLEF-DA}\footnote{http://imageclef.org/2014/adaptation} and \emph{Digits-five} have been introduced for the MDA experimental evaluation. \emph{Office-31} is a object recognition benchmark with 31 categories and 4652 images unevenly spread in three visual domains \textbf{A} (\emph{Amazon}), \textbf{D} (\emph{DSLR}), \textbf{W} (\emph{Webcam}). \emph{ImageCLEF-DA} derives from ImageCLEF 2014 domain adaptation challenge, and is organized by selecting 12 object categories (aeroplane, bike, bird, boat, bottle, bus, car, dog, horse, monitor, motorbike, and people) shared in the three famous real-world datasets, \textbf{I} (\emph{ImageNet ILSVRC 2012}), \textbf{P} (\emph{Pascal VOC} 2012), \textbf{C} (\emph{Caltech-256}). It includes 50 images in each category and totally 600 images for each domain. \emph{Digits-five} includes five digit image sets respectively sampled from following public datasets, \textbf{mt} (\emph{MNIST}) \cite{lecun1998gradient}, \textbf{mm} (\emph{MNIST-M}) \cite{Ganin2017Domain}, \textbf{sv}(\emph{SVHN}) \cite{Netzer2011Reading}, \textbf{up} (\emph{USPS}) and \textbf{sy} (\emph{Synthetic Digits}) \cite{Ganin2017Domain}. Towards the images in \emph{MNIST}, \emph{MNIST-M}, \emph{SVHN} and \emph{Synthetic Digits}, we draw 25000 for training and 9000 for testing in each dataset. There are only 9298 images in \emph{USPS}, so we choose the entire dataset as our domain.


\subsection{Evaluations in the vanilla setting}

\begin{table}[htp!]
	\center
	\caption{ Classification accuracy (\%) on Office-31 dataset for MDA in the vanilla setting. }\label{t1}
	\begin{footnotesize}
		\begin{tabular}{|l|c|ccc|c|cccr}
			\thickhline
			Standards &Models &\begin{tiny}
				A,W$\rightarrow$D
			\end{tiny} &\begin{tiny}A,D$\rightarrow$W\end{tiny} &\begin{tiny}D,W$\rightarrow$A\end{tiny} &Avg\\
			\hline
			\multirow{5}{0.3cm}{Single best}					&TCA		&95.2		&93.2		&51.6	 &68.8 		\\
			&GFK				&95.0		&95.6	  	&52.4	&68.7 		\\
			&DDC		&98.5		&95.0		&52.2	  &70.7		\\
			&DRCN		&99.0		&96.4		&\textbf{56.0} &73.6 		\\
			&RevGrad				&99.2		&96.4	  	&53.4	&74.3 		\\
			&DAN				&99.0		&96.0	  	&54.0		 &72.9		\\
			&RTN				&\textbf{99.6}		&96.8	  	&51.0		 &73.7		\\
			\hline
			\multirow{3}{0.3cm}{Source combine}		&Source only			&98.1		&93.2		&50.2	  &80.5		\\
				&RevGrad			&98.8		&96.2		&54.6	 &83.2 		\\
			&DAN 			&98.8		&95.2		&53.4	 &82.5 		\\
			\hline													\multirow{4}{0.3cm}{Multi-source}
			&Source only & 98.2  &92.7 &51.6 &80.8\\
			&sFRAME 		&54.5		&52.2	&32.1		&46.3	  		\\
			&SGF		&39.0		&52.0		&28.0		&39.7	  		\\				
			&DCTN (ours)	&\textbf{99.6}		&\textbf{96.9}		&54.9 			  	&\textbf{83.8}	\\
			\thickhline
		\end{tabular}
	\end{footnotesize}
	
	\caption{ Classification accuracy (\%) on ImageCLEF-DA dataset for MDA in the vanilla setting. }\label{t2}
	\begin{footnotesize}
		\begin{tabular}{|l|c|ccc|c|cccr}
			\thickhline
			Standards &Models &\begin{tiny}I,C$\rightarrow$P\end{tiny} &\begin{tiny}I,P$\rightarrow$C\end{tiny} &\begin{tiny}P,C$\rightarrow$I\end{tiny} &Avg\\
			\hline
			\multirow{3}{0.3cm}{Single best}
			&RevGrad				&66.5		&89.0	  	&81.8	&78.2 		\\					
			&DAN			&67.3		&88.4	  	&80.5	 &76.9		\\
			&RTN				&67.4		&89.5	  	&81.3 &78.4	 		\\
			
			\hline
			\multirow{3}{0.3cm}{Source combine}	&Source only		&68.3	&88.0	&81.2	 &79.2 		\\
						&RevGrad 			&67.0		&90.7		&81.8	 &79.8 		\\
			&DAN 		&\textbf{68.8}		&88.8		&81.3	&79.6  		\\
			
			\hline													\multirow{2}{0.3cm}{Multi-source}
			&Source only & 68.5  &89.3 &81.3 &79.7\\				
			&DCTN (ours)	&\textbf{68.8}		&\textbf{90.0}		&\textbf{83.5}	&\textbf{80.8}		  		\\
			\thickhline
		\end{tabular}
	\end{footnotesize}
	
	\caption{ Classification accuracy (\%) on Digits-five dataset for MDA in the vanilla setting.  }\label{t3}
	\begin{footnotesize}
		\begin{tabular}{|l|c|cc|c|cccccccr}
			\thickhline
			\multirow{2}{1cm}{Standards} &\multirow{2}{1cm}{Models} &\multirow{2}{1.5cm}{mm, mt, sy, up $\rightarrow$ sv} &\multirow{2}{1.5cm}{mt, sy, up, sv $\rightarrow$ mm} &Avg \\
			
			& & & & \\ \hline
			\multirow{3}{0.3cm}{Source combine}	&Source only					&72.2		&64.1	 &68.2 		 		\\
						&RevGrad 			&68.9		&\textbf{71.6}		&70.3 		\\
			&DAN	&71.0	&66.6	&	68.8  		\\
			\hline												\multirow{4}{0.3cm}{Multi-source}
			
			&Source only				&64.6		&60.7 &62.7	\\
						&RevGrad 			&61.4		&71.1		&66.3 		\\
			&DAN				&62.9		&62.6 &62.8 \\			
			&DCTN (ours)	&\textbf{77.5}		&70.9	&\textbf{74.2}				  		\\
			\thickhline\thickhline
		\end{tabular}
	\end{footnotesize}
\end{table}

\textbf{Baselines.} The existing works of MDA lack comprehensive evaluations on real-world visual recognition benchmarks. In our experiment, we introduce two shallow methods, sparse FRAME (\textbf{sFRAME}) \cite{xie2015learning} and \textbf{SGF} \cite{gopalan2011domain} as the multi-source baselines in the \emph{Office-31} experiment. Besides, we evaluate DCTN with single-source visual UDA methods including the conventional, e.g., Transfer Component Analysis (\textbf{TCA}) \cite{pan2011domain} and Geodesic Flow Kernel (\textbf{GFK}) \cite{gong2012geodesic}, as well as state-of-the-art deep methods: Deep Domain Confusion (\textbf{DDC}) \cite{hoffman2017simultaneous}, Deep Reconstruction-classification Networks
(\textbf{DRCN}) \cite{ghifary2016deep}, Reversed Gradient (\textbf{RevGrad}) \cite{ganin2015unsupervised}, Domain Adaptation Network (\textbf{DAN}) \cite{long2015learning}, and Residual Transfer Network (\textbf{RTN}) \cite{long2016unsupervised}. Since those methods perform in single-source setting, we introduce two MDA standards for different purposes: 1). \emph{Source combine}: all source domains are combined into a traditional single-source \emph{v.s.} target setting. 2). \emph{Single best}: in the muli-source domains, we report the single source transfer result best-performing in the test set. The first standard testify whether the multi-source is valuable to exploit; the second evaluates whether we can further improve the best single source UDA via introducing another source transfer. Additionaly, as baselines in the \emph{Source combine} and multi-source standards, we use all images from sources to train backbone-based multi-source classifiers and directly apply them to classify target images. They are termed \emph{Source only} and used to confirm whether our multi-source transfers are available. For a fair comparison, all deep model baselines in \emph{Office-31} and \emph{ImageCLEF-DA} use the Alexnet architectures, and share the same backbone model in \emph{Digits-five}.

\begin{table}[htp!]
	\center
	\caption{ Evaluations on Office-31 (A,D$\rightarrow$ W) for MDA in the category shift setting.}\label{t4}
	\begin{footnotesize}
		\begin{tabular}{|l|cccc|ccccr}
			\thickhline
			
			\multirow{2}{1.2cm}{Category Shift} &\multirow{2}{0.8cm}{Models}
			&\multirow{2}{1cm}{Accuracy} &\multirow{2}{1.cm}{Degraded Accuracy}
			&\multirow{2}{1cm}{Transfer Gain}  \\
			
			& & &  \\ \hline
			\multirow{3}{0.6cm}{Overlap}
			&Source only			&84.4		&-8.3	  	&0 		\\					
						&RevGrad 			&86.3		&-7.9		&1.9	 		\\
			&DAN					&87.8		&	 -\textbf{6.4} & 	 3.4		\\
			&DCTN(ours)					&90.2		&-6.7	  	&\textbf{5.8}	 		\\\hline	
			\multirow{3}{0.6cm}{Disjoint}
			&Source only			&78.1		&-14.6	 &0 \\
						&RevGrad 			&78.6			&-15.6	 &0.5 		\\		&DAN				&75.5		&	  	-18.7 & -2.6		\\
			&DCTN(ours)					&82.9		&\textbf{-14.0}	  			&\textbf{4.8} 		\\
			\thickhline
		\end{tabular}
	\end{footnotesize}
	
	\caption{  Evaluations on ImageCLEF-DA (I,P$\rightarrow$ C) for MDA in the category shift settings.  }\label{t5}
	\begin{footnotesize}
		\begin{tabular}{|l|cccc|ccccr}
			\thickhline			
			\multirow{2}{1.2cm}{Category Shift} &\multirow{2}{0.8cm}{Models}
			
			&\multirow{2}{1cm}{Accuracy} &\multirow{2}{1cm}{Degraded Accuracy}
			&\multirow{2}{1cm}{Transfer Gain}  \\
			
			& & &  \\ \hline
			\multirow{3}{0.6cm}{Overlap}	
			&Source only				&86.3		&-3.0	  	&0	 		\\		
						&RevGrad 			&85.7		&-4.5		 &-0.6 		\\		
			&DAN			&85.5		&-4.0 &-0.8	  	 		\\
			&DCTN(ours)			 &88.7		&\textbf{-1.3}		  	 	&\textbf{2.4}		\\
			\hline	
			\multirow{3}{0.6cm}{Disjoint}			&Source only					&81.5		&\textbf{-7.8}	  	&0	 		\\
						&RevGrad 			&71.5		&-18.7			 & -10.0		\\	&DAN				&71.0		&-18.5 &-10.5	  	 		\\
			&DCTN(ours)			&82.0		&-8.0	  	&\textbf{0.5}	 		\\
			\thickhline
		\end{tabular}
	\end{footnotesize}
\end{table}

In the object recognition, we report all combinations of domain shifts and compare DCTN with the baselines. Tables.\ref{t1}-\ref{t2} show that DCTN yields the best results in the \emph{Office-31} transfer tasks \textbf{A},\textbf{W}$\rightarrow$\textbf{D} and \textbf{A},\textbf{D}$\rightarrow$\textbf{W}, performs compelling in \textbf{D},\textbf{W}$\rightarrow$ \textbf{A} and outperforms conventional MDA baselines by large margins. In the \emph{ImageCLEF-DA}, DCTN attains the state of the art in all transfer tasks. These validate that, no matter domain size is equal or not, DCTN can learn more transferable and discriminative features from multi-source transfer. 


In the digit recognition, there are four source domains and we convey the results in the domain shifts as \textbf{mm, mt, sy, up} $\rightarrow$ \textbf{sv} and \textbf{mt, sv, sy, up} $\rightarrow$ \textbf{mm}. We compare them with DAN under the source-combine and the multi-source average accuracy of its four single source transfer combinations. The results have been shown in Table.\ref{t3}. Despite of involving multiple source domain shifts, DCTN still can improve the source combine performance by $6.0\%$. 

\begin{figure*}[!hbt]
	\centering
	\includegraphics[width = 6in]{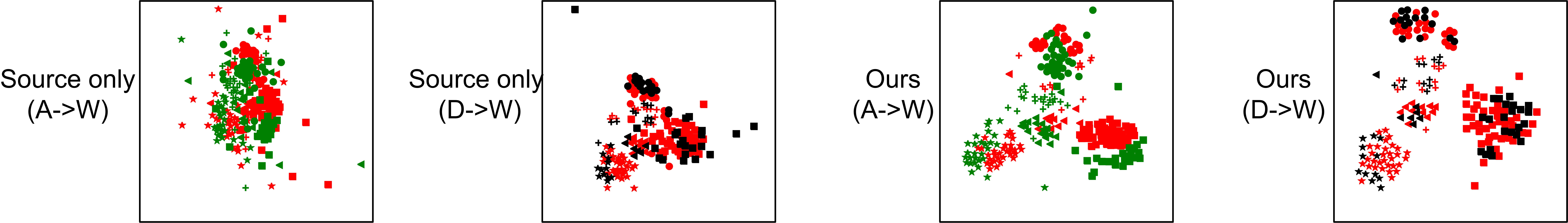}
	\caption{The t-SNE \cite{maaten2008visualizing} visulization of A,D$\rightarrow$W. Green, black and red represent domains A, D and W respectively. We use different markers to denote 5 categories, e.g., bookcase, calculator, monitor, printer, ruler. Best viewed in color.} \label{v1}
	\centering
	\includegraphics[width = 6in]{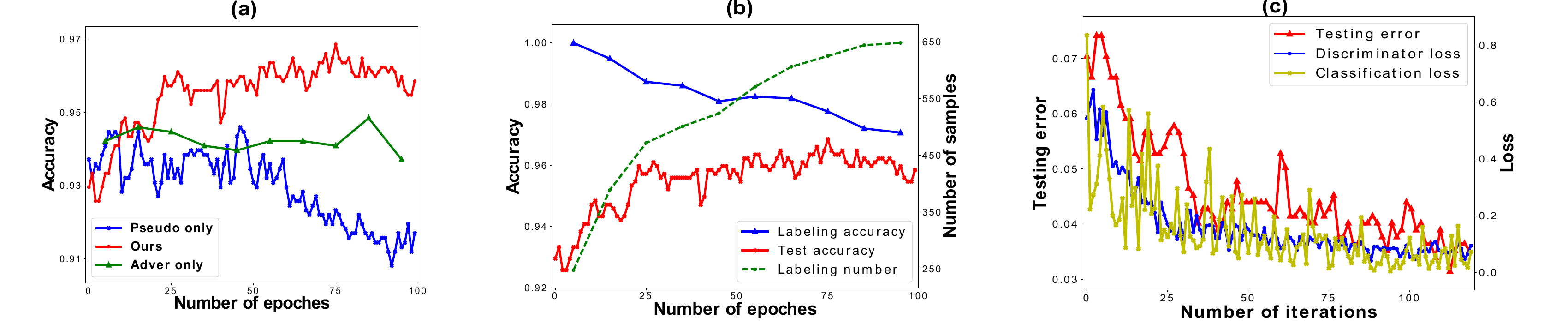}
	\caption{Analysis:(a) the accuracies of DCTN, adversarial-only and pseudo-only models; (b) the accuracies of testing samples and pseudo-labeled target samples; (c) the convergence performance on different losses. Best viewed in color. }\label{v2}
\end{figure*}

\subsection{Evaluations in the category shift setting}

\textbf{How to evaluate?} Since category shift is a brand new MDA protocol, in order to evaluate the model in  this protocol, the multiple sources are amended to satisfy categorical disalignments. We consider the two-source adaptation in object recognition. In the category order of the benchmarks, we take the first and the last one third classes as the private classes of both source domains respectively, and the rest are the public classes shared in both sources. This organization in category shift is termed \emph{Overlap}. In the same order, we depart all categories into two non-overlapped class sets and define them as the private classes. Since no classes are common, we named it as \emph{Disjoint}. We testify DCTN on both the source domain organizations, and compare the results with Source only, RevGrad and DAN. The accuracy degradation compared to the performance in the vanilla setting and the transfer gain compared to \emph{Source only} are also appended.

The evaluations have been shown in Table.\ref{t4}-\ref{t5}. Category shift is very challenging, and under the \emph{Overlap}, the accuracies of DAN got slashed by $-6.4$ in the \emph{Office-31} and $-4.0$ in the \emph{ImageCLEF-DA}. The performance deteriorate to $-18.7$ and $-18.5$ under the \emph{Disjoint}. Moreover, DAN also suffers negative transfer gains in most situations, which indicates the transferbility of DAN cripled in the category shift. In contrast, DCTN reduces the performance drops compared to the model in the vanilla setting, and obtains positive transfer gains in all situations. It reveals that DCTN can resist the negative transfer caused by the category shift.


\subsection{Further Analysis}

\textbf{Feature visualization.} In the experiment of adaptation task \textbf{A,D} $\rightarrow$ \textbf{W} in Office-31, we visualize the DCTN activations before and after adaptation. For simplicity, both the source domains have been separated to emphasize the contrast of target. As we can see in Fig.\ref{v1}, compared with the activations given by the source only, both of the activations from \textbf{A} $\rightarrow$ \textbf{W} and \textbf{D} $\rightarrow$ \textbf{W} have shown good adaptation patterns. It means DCTN can successfully learn transferable features with multiple sources. Besides, the target activations become more clear to categorize, which suggests that the features learned by DCTN attains desirable discriminative property. Finally, even if the multi-source transfer has been composed of hard transfer task ( \textbf{A} $\rightarrow$ \textbf{W} ), DCTN is still able to adapt to target domain without the degradation in the performance of \textbf{D} $\rightarrow$ \textbf{W}.

\textbf{Ablation study.} DCTN contains two major parts: the multi-way adversary and the auto-labeling scheme. To further reveal their function, we decompose DCTN into two variants: The \textbf{adversarial-only} model excludes the pseudo-labels and updates the category classifier with source samples. The \textbf{pseudo-only} model forbids the adversary and categorize target samples with average multi-source results. As shown in Fig.\ref{v2}(a), the accuracy of adversary behaves stable in each iteration, but lack of target class guidance, its final performance hits a bottleneck. But without the adversary, the accuracy of pseudo labels significantly drops and pulls down the DCTN accuracy. It indicates that the both adaptations cooperate with each other to achieve desirable transfer behaviors. Diving deeper in Fig.\ref{v2}(b), the test accuracy and the pseudo label accuracy show converged in the alternative learning, which implicitly reveals the consistency between the both adaptations. We also provide the ablative study result to the domain batch mining technique (see Table.\ref{t6}), which testify the method's efficacy.  
\begin{table}
	\center
	\caption{Ablation study of Algorithm.\ref{A2} in Office-31. }\label{t6}
	\begin{footnotesize}
		\begin{tabular}{|l|ccc|cc|ccr}
			\thickhline			
			&A,W$\rightarrow$D	&A,D$\rightarrow$W		&D,W$\rightarrow$A		&Overlap	  	& Disjoint		\\	
			\hline				
			w&\textbf{99.6}&\textbf{96.9}&54.9&\textbf{90.2}&\textbf{82.9} \\
			w/o	&99.0&96.1&\textbf{55.0}&89.3&82.6 \\
			\thickhline
		\end{tabular}
	\end{footnotesize}
\end{table}

\textbf{Convergence analysis.} As DCTN involves a complex learning procedure including adversarial learning and alternative adaptation, we testify the convergence performance of different losses. During the process of hard sub transfer A$\rightarrow$W, Fig.\ref{v2}(c) demonstrates that, despite of the frequent deviation, the classification loss, adversarial loss and testing error gradually converge.

\section{Conclusion}

In this paper we have explored the unsupervised domain adaptation with multiple source domains. We raise a new MDA protocol termed \emph{category shift}, where classes from different sources are non-consistent. Furthermore, we proposed \emph{deep cocktail network}, a novel framework to obtain transferable and discriminative features from multiple sources. The approach can be applied to the ordinary MDA setting and category shift, and more, achieves state-of-the-art results in most of our evaluation protocols.


{\small
	\bibliographystyle{ieee}
	\bibliography{reference}
}

\end{document}